\pdfoutput=1

\documentclass[11pt]{article}

\usepackage[final]{arxiv}

\usepackage{times}
\usepackage{latexsym}

\usepackage[T1]{fontenc}
\usepackage[utf8]{inputenc}
\usepackage{microtype}
\usepackage{inconsolata}

\usepackage{hyperref}       
\usepackage{url}            
\usepackage{booktabs}       
\usepackage{amsfonts}       
\usepackage{nicefrac}       
\usepackage{microtype}      
\usepackage{xcolor}         

\usepackage{algorithm}
\usepackage{enumitem}
\usepackage{amsmath}
\usepackage{mathrsfs}
\usepackage{amssymb}
\usepackage{subfigure}
\usepackage{graphicx}
\usepackage{amsthm}
\usepackage{color}
\usepackage{mathrsfs}
\usepackage{amsfonts}
\usepackage{wrapfig}
\usepackage{hyperref}
\usepackage{blindtext}
\usepackage{caption}
\usepackage{makecell}
\usepackage{algpseudocode}
\usepackage{arydshln}
\usepackage{tabularx}

\usepackage{pifont}
\def\Algname{Multi-Agent Filtering RAG}

\def\Algnameabbr{\texttt{MAIN-RAG}}

\definecolor{darkred}{RGB}{255,0,0}
\definecolor{darkblue}{RGB}{0,0,180}
\definecolor{darkgreen}{HTML}{00a64f}

\usepackage[textsize=small]{todonotes}

\algnewcommand\algorithmicinput{\textbf{Input:}}
\algnewcommand\INPUT{\item[\algorithmicinput]}
\algnewcommand\algorithmicoutput{\textbf{Output:}}
\algnewcommand\OUTPUT{\item[\algorithmicoutput]}

\title{\Algnameabbr{}: Multi-Agent Filtering Retrieval-Augmented Generation}

\author{
  \small Chia-Yuan Chang \\
  \small Texas A\&M University \\
  \small \texttt{cychang@tamu.edu}\\ 
  \And
  \small Zhimeng Jiang \\
  \small Visa Research \\
  \small \texttt{zhimjian@visa.com}\\
  \And
  \small Vineeth Rakesh \\
  \small Visa Research \\
  \small \texttt{vinemoha@visa.com}\\
  \And
  \small Menghai Pan \\
  \small Visa Research \\
  \small \texttt{mengpan@visa.com}\\
  \And
  \small Chin-Chia Michael Yeh \\
  \small Visa Research \\
  \small \texttt{miyeh@visa.com}\\
  \AND
  \small Guanchu Wang \\
  \small Visa Research \\
  \small \texttt{guanwang@visa.com}\\
  \And
  \small Mingzhi Hu \\
  \tiny Worcester Polytechnic Institute \\
  \small \texttt{mhu3@wpi.edu}\\
  \And
  \small Zhichao Xu \\
  \small University of Utah \\
  \small \texttt{zhichao.xu@utah.edu}\\
  \And
  \small Yan Zheng \\
  \small Visa Research \\
  \small \texttt{yazheng@visa.com}\\
  \And
  \small Mahashweta Das \\
  \small Visa Research \\
  \small \texttt{mahadas@visa.com}\\
  \And
  \small Na Zou \\
  \small University of Houston \\
  \small \texttt{nzou2@central.uh.edu}\\ 
  }

\begin{document}
\maketitle
\begin{abstract}

Large Language Models (LLMs) are becoming essential tools for various natural language processing tasks but often suffer from generating outdated or incorrect information. Retrieval-Augmented Generation (RAG) addresses this issue by incorporating external, real-time information retrieval to ground LLM responses. However, the existing RAG systems frequently struggle with the quality of retrieval documents, as irrelevant or noisy documents degrade performance, increase computational overhead, and undermine response reliability.
To tackle this problem, we propose Multi-Agent Filtering Retrieval-Augmented Generation (\Algnameabbr{}), a training-free RAG framework that leverages multiple LLM agents to collaboratively filter and score retrieved documents. Specifically, \Algnameabbr{} introduces an adaptive filtering mechanism that dynamically adjusts the relevance filtering threshold based on score distributions, effectively minimizing noise while maintaining high recall of relevant documents. The proposed approach leverages inter-agent consensus to ensure robust document selection without requiring additional training data or fine-tuning.
Experimental results across four QA benchmarks demonstrate that \Algnameabbr{} consistently outperforms traditional RAG approaches, achieving a 2–11\% improvement in answer accuracy while reducing the number of irrelevant retrieved documents. Quantitative analysis further reveals that our approach achieves superior response consistency and answer accuracy over baseline methods, offering a competitive and practical alternative to training-based solutions.

\end{abstract}

\section{Introduction}
\label{sec:intro}

Large Language Models (LLMs) have revolutionized natural language processing (NLP), achieving state-of-the-art performance across diverse tasks, such as question answering, summarization, and text generation~\cite{vaswani2017attention, brown2020language}. However, their reliance on pre-trained static data introduces critical limitations, including the generation of outdated or factually incorrect information—a phenomenon referred to \textit{hallucination}~\cite{ji2023survey, zhang2023siren}. This issue becomes particularly pronounced in applications requiring accurate, up-to-date, and contextually relevant responses, such as healthcare, legal systems, and real-time customer support~\cite{bommasani2021opportunities, zellers2019defending, lin2021truthfulqa}.

Retrieval-augmented generation (RAG) has emerged as a promising solution to mitigate these challenges by integrating real-time document retrieval to ground LLM outputs in reliable external knowledge~\cite{lewis2020retrieval, guu2020retrieval, karpukhin2020dense, ram2023context, li2023llatrieval, wang2023self}. 
Training-based methods~\cite{guu2020retrieval, karpukhin2020dense, wang2023self} have demonstrated strong performance but require substantial computational resources and training time. 
In contrast, training-free approaches~\cite{ram2023context, li2023llatrieval, jiang-etal-2023-active}, while simpler and more efficient as plug-and-play methods, still hinge on the quality of retrieved documents~\cite{chen2024benchmarking, yu2024rankrag}. The presence of irrelevant or noisy documents not only reduces response accuracy but also increases computational overhead and compromises system reliability. These challenges underscore the urgent need for robust mechanisms to effectively \textit{filter} and\textit{rank} retrieved content.

\begin{figure*}[t!]
\centerline{\includegraphics[width=1.\textwidth]{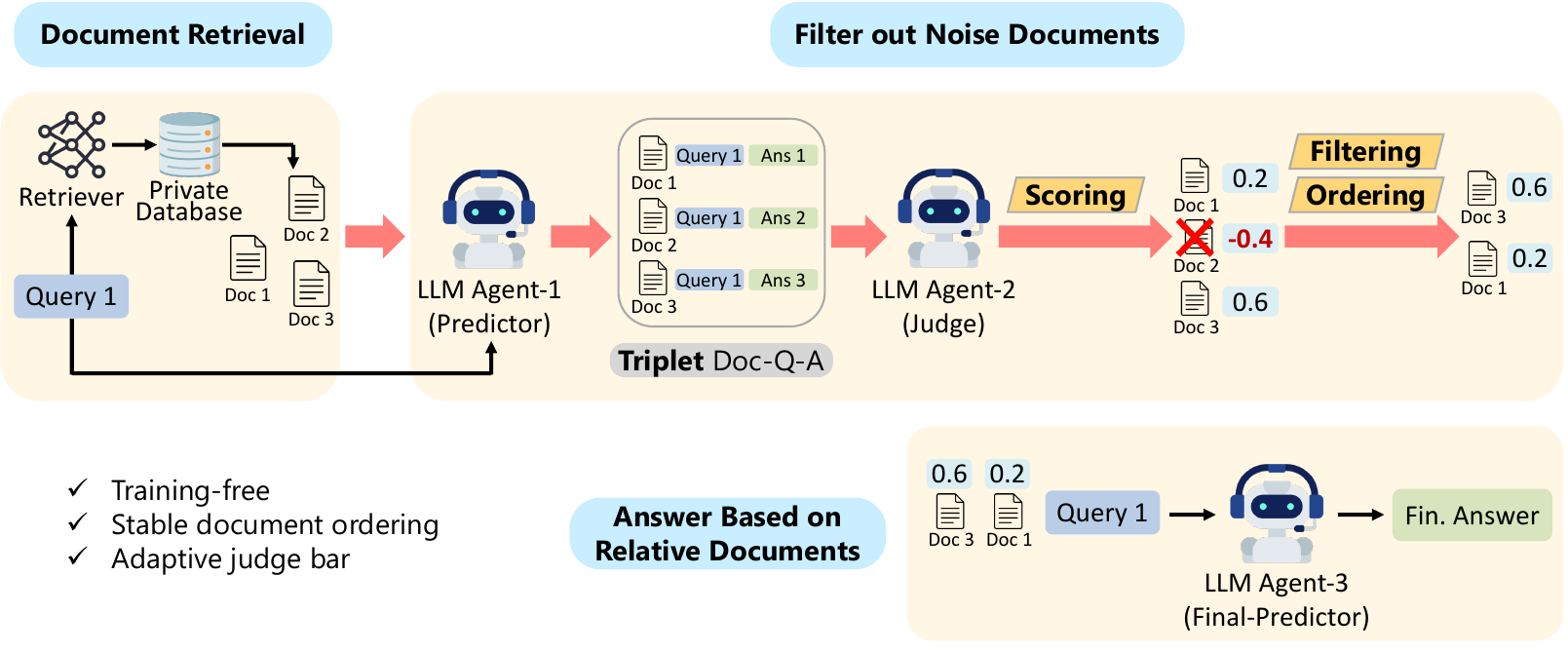}}
\caption{An overview of the proposed framework \Algnameabbr{}, consisting of three LLM agents to identify noisy retrieved documents for filtering (see Section~\ref{sec:define_agents}). After the retrieval, \texttt{Agent-1 "Predictor"} infers answers for each query; then, \texttt{Agent-2 "Judge"} takes \textit{Doc-Q-A Triplet} to judge if a document is supportive for LLMs to answer the query. \texttt{"Judge"} provides relevant scores for each document for filtering and ordering. Finally, \texttt{Agent-3 "Final-Predictor"} answers the query with the given document list.}
\label{fig:model_layout}
\end{figure*}

To address these challenges, we propose \textit{\textcolor{darkred}{M}ulti-\textcolor{darkred}{A}gent F\textcolor{darkred}{I}lteri\textcolor{darkred}{N}g \textcolor{darkred}{R}etrieval-\textcolor{darkred}{A}ugmented \textcolor{darkred}{G}eneration} (\Algnameabbr{}), a novel training-free framework designed to enhance the performance and reliability of RAG systems. Unlike existing methods that often rely on computationally intensive training or fine-tuning, \Algnameabbr{} leverages a collaborative multi-agent approach where multiple LLM agents filter and score retrieved documents. This consensus-driven strategy ensures that only the most relevant and high-quality documents are utilized for generation, significantly reducing noise without sacrificing recall.

\Algnameabbr{} introduces an adaptive filtering mechanism that dynamically adjusts the relevance threshold based on the score distribution of retrieved documents. This adaptability allows the framework to handle diverse queries effectively and ensures robust performance across diverse tasks. Furthermore, the training-free nature of \Algnameabbr{} eliminates the need for additional labeled data or model fine-tuning, making it a scalable and versatile solution for real-world applications.

Our contributions are as follows:

\begin{itemize}
    \item \textbf{Training-Free Multi-Agent Filtering}: We introduce a novel training-free framework that employs multiple LLM agents to collaboratively filter and rank retrieved documents, improving retrieval precision and RAG reliability without the need for additional training.
    \item \textbf{Dynamic and Adaptive Filtering Mechanism}: \Algnameabbr{} incorporates an adaptive threshold mechanism that dynamically adjusts to query-specific score distributions, ensuring effective noise reduction while maintaining high recall of relevant documents.
    \item \textbf{Empirical Validation Across Multiple Benchmarks}: Our experimental results on four QA benchmarks demonstrate that \Algnameabbr{} outperforms baseline RAG approaches, achieving a 2-11\% improvement in answer accuracy while reducing the inclusion of irrelevant documents.
\end{itemize}

By addressing the inherent challenges of noise in document retrieval and providing a training-free solution, \Algnameabbr{} represents a significant advancement in the field of retrieval-augmented generation. This work details the design, implementation, and evaluation of \Algnameabbr{}, highlighting its potential to improve response accuracy, consistency, and reliability in diverse NLP applications.

\section{Preliminaries}

\subsection{Notations and Objectives}
We consider an RAG system designed to filter noisy retrieved documents and improve response accuracy. Each query $q \in \mathcal{Q}$ retrieves a set of documents $\mathcal{D}_q = \{d_1, d_2, \dots, d_N\}$ using a retriever model. Each document $d_i$ is associated with a relevance score $r_i$, which quantifies its usefulness for the query and is determined by \textbf{\texttt{Agent-2 (Judge)}} as described in Section~\ref{sec:judgement_quant}.
Let $\boldsymbol{R} = [r_1, r_2, \dots, r_N]$ represent the relevance scores for the retrieved documents. These scores are used to rank the documents, forming an ordered list $\mathcal{D}_q^{\text{rank}}$, where documents with higher scores are deemed more relevant. Based on these scores, an adaptive judge bar $\tau_q$ is computed for each query to filter out noisy documents (see Section~\ref{sec:adaptive_bar}). Documents with scores $r_i \geq \tau_q$ are retained, creating a filtered set $\mathcal{D}_q^{\text{filtered}} \subseteq \mathcal{D}_q^{\text{rank}}$.
For $1 \leq i \leq N$, $r_i$ represents the relevance score for document $d_i$. The adaptive judge bar $\tau_q$ dynamically adjusts based on the distribution of $\boldsymbol{R}$, ensuring robust filtering for diverse queries.
For example, consider a query $q$ that retrieves $\mathcal{D}_q = \{d_1, d_2, d_3\}$ with relevance scores $\boldsymbol{R} = [3.8, 2.5, 4.2]$. The ranked list $\mathcal{D}_q^{\text{rank}}$ becomes $\{d_3, d_1, d_2\}$. If the adaptive judge bar $\tau_q = 3.0$, the filtered set $\mathcal{D}_q^{\text{filtered}} = \{d_3, d_1\}$ retains only the most relevant documents.
To this end, our work focuses on effectively identifying and filtering noisy documents, thereby enhancing the accuracy and reliability of RAG systems in a post-hoc manner.

\subsection{Impact of Noisy Retrieval Documents}
In RAG, irrelevant or noisy documents retrieved during the retrieval stage can mislead the LLMs during the inference stage, often resulting in incorrect answers. 
The presence of such noise information poses a significant challenge to the reliability of LLMs and RAG, especially when applied to tasks that require precise information, such as question answering. 
As observed in existing studies~\cite{chen2024benchmarking, yu2024rankrag}, LLMs exhibit vulnerabilities in noise robustness and often fail to reject irrelevant content, resulting in decreased performance. 
Therefore, improving noise filtering after the retrieval process is vital to enhance RAG systems' reliability and robustness.

\subsection{Related Works}

This section reviews RAG methodologies, focusing on training-based and training-free approaches, and discusses the challenge of noise robustness in RAG.

\noindent \textbf{Training-based RAG.}
Training-based RAG integrates retrieval mechanisms into the training of the language model, allowing access to external information during generation. For instance, \citet{lewis2020retrieval} combine parametric and nonparametric pre-trained memory for language generation, achieving state-of-the-art results on open-domain QA tasks. Similarly, \citet{guu2020retrieval} introduce REALM, a framework that augments language model pre-training with a latent knowledge retriever, allowing retrieval and attention to large corpora like Wikipedia. Self-RAG~\cite{asaiself} proposes to adaptively retrieve passages and critique the generations so as to improve output quality and factuality.
Albeit effective, these methods require dedicated training procedures and corresponding hardware, hindering their applicability.

\noindent \textbf{Training-free RAG.}
Training-free RAG approaches integrate pre-trained language models with retrieval components, avoiding extensive retraining.~\citet{ram2023context} perform in-context retrieval, allowing language models to dynamically access external data.~\citet{li2023llatrieval} propose a framework where LLMs verify retrieved documents to ensure their relevance to queries, but this method is highly sensitive to input prompts. Similarly,~\citet{jiang-etal-2023-active} introduce a strategy to actively determine when and what to retrieve during generation, but it also suffers from prompt sensitivity. While efficient, training-free RAG approaches struggle with noise robustness due to their reliance on static pre-trained data.

\noindent \textbf{Challenge of noise robustness in RAG.}
Ensuring noise robustness is critical for the reliability of RAG systems.~\citet{chen2024benchmarking} conduct a comprehensive analysis of RAG's effects on LLMs, focusing on their resilience to noise and other fundamental capabilities.~\citet{yu2024rankrag} present a framework that strengthens LLMs' RAG performance by guiding them in context ranking and answer generation. Section 3.2, "Trade-off of Picking Top-k Contexts," underscores the significance of selecting relevant contexts to balance effectiveness and computational cost. These findings emphasize the necessity of filtering out noisy documents to uphold the accuracy and robustness of RAG systems.

\section{\Algname{} (\Algnameabbr{})}   
\label{sec:method}
This section presents a comprehensive overview of our proposed \Algnameabbr{} framework, as depicted in Figure~\ref{fig:model_layout}.
Based on the traditional RAG workflow, \Algnameabbr{} focuses on reducing noisy documents after the retrieval stage.
Specifically, \Algnameabbr{} is a training-free framework, involving three agents to identify and filter out noisy documents after retrieval. 
The specific roles of the three agents are defined in Section~\ref{sec:define_agents}.
Section~\ref{sec:judgement_quant} illustrates the process of supportive document judgment for filtering out misleading or irrelevant ones.
Section~\ref{sec:adaptive_bar} proposes an adaptive judge bar to adjust the judge criteria according to given retrieved documents.

\begin{figure}[tb]
\centerline{\includegraphics[width=.5\textwidth]{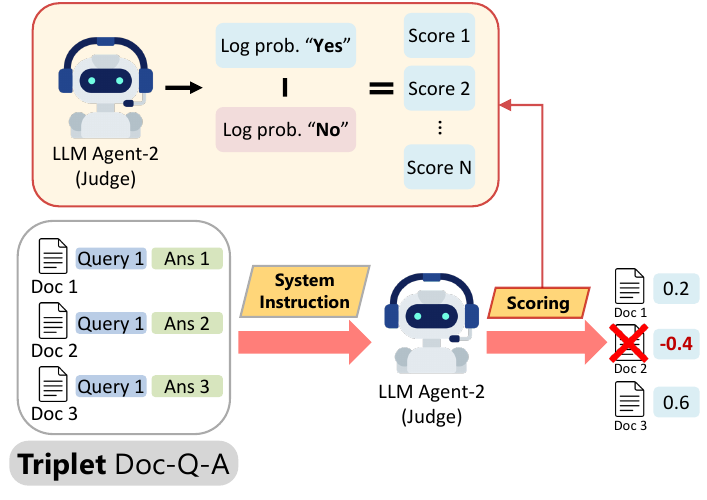}}
\caption{Quantification of document relevant score.}
\label{fig:quant_relvant_score}
\end{figure}

\subsection{Definition of LLM Agents in \Algnameabbr{}}
\label{sec:define_agents}
The proposed framework \Algnameabbr{} to identify noisy retrieved documents for filtering out, consisting of three LLM agents: \textbf{\texttt{Agent-1 (Predictor)}}, \textbf{\texttt{Agent-2 (Judge)}}, and \textbf{\texttt{Agent-3 (Final-Predictor)}}.

\textbf{\texttt{Agent-1 (Predictor)}.}
After the document retrieval stage, we have several candidate documents for each query.
Then, for a single query, \texttt{Agent-1 (Predictor)} is to infer answers to the query given each document. Then, we can form the Document-Query-Answer Triplet (\textbf{Doc-Q-A}), which is prepared for \texttt{Agent-2 (Judge)} to evaluate the relevant information among Doc-Q-A triplet, as shown in Figure~\ref{fig:model_layout}.

\textbf{\texttt{Agent-2 (Judge)}.}
Given a Doc-Q-A triplet, \texttt{Agent-2 (Judge)} is to evaluate whether the document provides relevant information to the query and answer.
\texttt{Agent-2} is prompted to answer "\texttt{Yes}" or "\texttt{No}" for each Doc-Q-A triplet, treating the relevance judgment as a True-or-False question. 
This simplification helps to further quantify the judgment as relevant scores of documents, which can be used for filtering and ordering. 
The details of \texttt{Agent-2} refer to Section~\ref{sec:judgement_quant} and Section~\ref{sec:adaptive_bar}. 

\textbf{\texttt{Agent-3 (Final-Predictor)}.}
After \texttt{Agent-2} filters out noisy documents and orders the remaining document list by their relevant scores, \texttt{Agent-3 (Final-Predictor)} is prompted to answer the query with the document list.

\subsection{Relevance Judgment Quantification}
\label{sec:judgement_quant}
\begin{figure}[tb]
\centerline{\includegraphics[width=.48\textwidth]{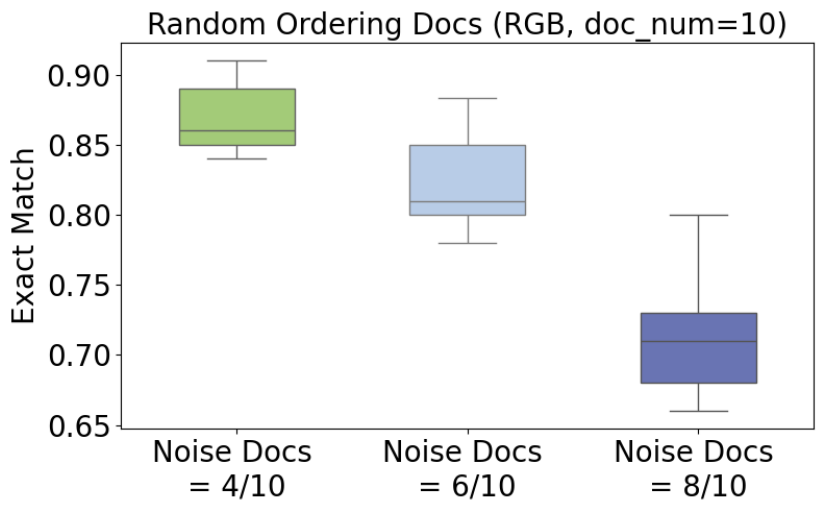}}
\caption{Impacts of document ordering on variance in RAG performance, where Noise Docs $t/u$ means $t$ noisy documents out of $u$ retrieved documents.}
\label{fig:exp_random_order}
\end{figure}

Previous research has observed that when processing long context inputs, LLMs tend to overlook information in the middle, placing greater emphasis on the beginning and end of the context~\cite{liu2024lost}. 
This suggests that in RAG, the ordering of documents may influence prediction performance.
To investigate the impact of document order in RAG, we conducted an experiment on the benchmark RGB~\cite{chen2024benchmarking}, where the retrieved documents were randomly shuffled and evaluated. 
This process was repeated ten times for each noise ratio condition. 
The results, illustrated in Figure~\ref{fig:exp_random_order}, reveal that document order has a significant effect on performance. 
Notably, the maximum performances are substantially higher than the minimum ones, suggesting that certain document orders can provide stable and optimal results. 
This observation leads us to propose a judgment quantification to make documents sortable.

To quantify the natural language outputs "Yes" and "No," we propose computing the difference between the log probabilities of the corresponding tokens, as shown in Figure~\ref{fig:quant_relvant_score}, where the system instruction is provided in Appendix~\ref{apx:agent_prompt}.
Since these are log probabilities, this difference effectively represents the logarithm of the ratio of the probabilities of the "Yes" token to the "No" token.
By subtracting the log probabilities of the two tokens, \texttt{Agent-2} simplifies the judgment by consolidating the two factors into a single score. 
This relevant score then serves as the sole criterion for document filtering.

\begin{figure}[b]
\centerline{\includegraphics[width=.48\textwidth]{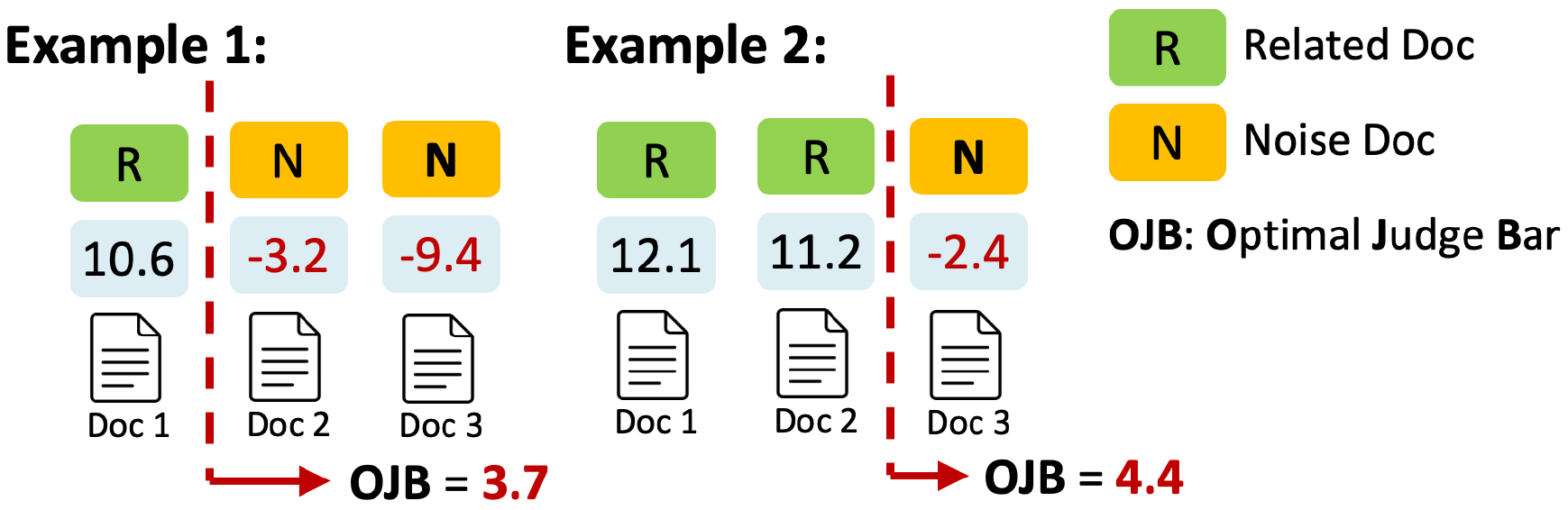}}
\caption{Examples of Optimal Judge Bar (\textbf{OJB}).}
\label{fig:optimal_bar_ex}
\end{figure}

\subsection{Adaptive Judge Bar $\tau_q$}
\label{sec:adaptive_bar}
\begin{figure}[tb]
\centerline{\includegraphics[width=.45\textwidth]{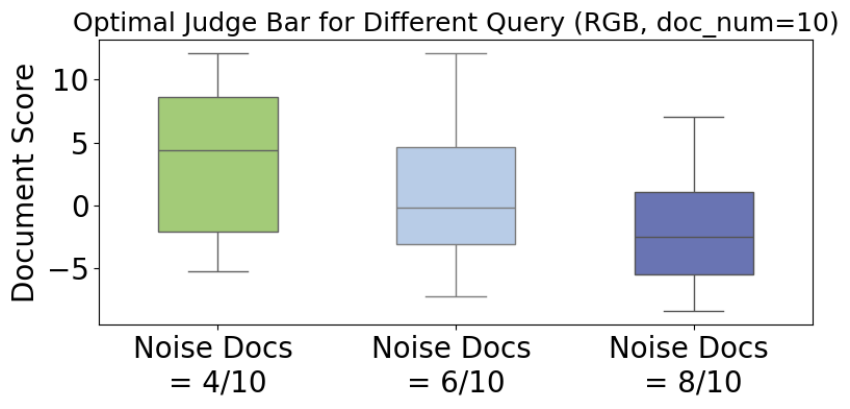}}
\caption{Optimal judge bars for different noise ratios in different queries, where Noise Docs $t/u$ means $t$ noisy documents out of $u$ retrieved documents.}
\label{fig:exp_optimal_bar}
\vspace{-0.5cm}
\end{figure}

After we obtain relevant scores for each document, another challenge is how to determine the optimal judge bar for filtering out noisy documents. 
Here, the optimal judge bar is the score that perfectly filters out all noisy documents while retaining all relevant ones.
Consider example 1 in Figure~\ref{fig:optimal_bar_ex}, where a query retrieves a higher number of noisy documents; the optimal judge bar in this case is approximately \texttt{3.7}. 
In example 2 in Figure~\ref{fig:optimal_bar_ex}, where more relevant documents are retrieved for a query, the optimal judge bar increases to around \texttt{4.4}. 
These examples illustrate that the optimal judge bar varies with the document distribution among queries.
We also observe significant variations in the optimal judge bars across different queries in RGB benchmark~\cite{chen2024benchmarking}, as shown in Figure~\ref{fig:exp_optimal_bar}.
This observation leads us to think about how can we adaptively determine optimal judge bars.

\begin{figure}[tb]
\centerline{\includegraphics[width=.5\textwidth]{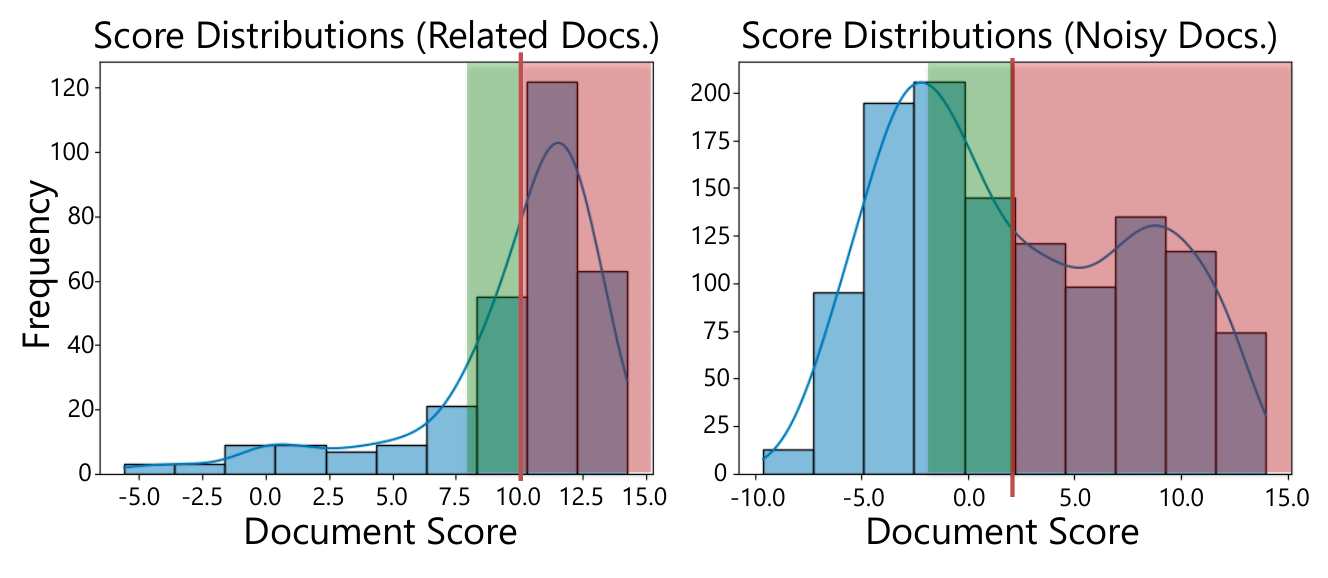}}
\caption{Score distribution of related and noisy documents with adaptive judge bar.}
\label{fig:score_dist}
\vspace{-0.3cm}
\end{figure}

Analyzing the relevant score distributions for both related and noisy documents on RGB benchmark~\cite{chen2024benchmarking}, we observe that the scores of related documents are skewed high with a small standard deviation, as shown in Figure~\ref{fig:score_dist}. 
This indicates that the LLM (here is Mistral-7B) is more confident about these documents. 
In contrast, the scores of noisy documents are more uniformly distributed with a larger standard deviation, suggesting that the LLM is less confident and may misjudge them.
Based on this biased LLM behavior, we propose using the average relevant score for each query as an adaptive judge bar. 
In Figure~\ref{fig:score_dist}, the red line represents the average score of all documents. 
Documents to the right of the red line (the red area) are retained, while those to the left are filtered out. 
When the average score is high—indicating many relevant documents—we can filter out most low-scoring outliers, which are likely noise. 
Conversely, when the average score is low—indicating many noisy documents—we aim to reduce the number of documents while maintaining a high recall rate for relevant documents by still using the average score to filter out approximately half of the documents.
To introduce flexibility into this framework, we adjust the adaptive judge bar $\tau_q$ by adding $n$ times the standard deviation $\sigma$ of each candidate document set, $\tau_q - n \cdot \sigma$, allowing relax $\tau_q$ when needed, as shown by the green area 
in Figure~\ref{fig:score_dist}. 
Notably, $n$ is the only hyperparameter in \Algnameabbr{}.

\begin{figure*}[tb]
\centerline{\includegraphics[width=1.\textwidth]{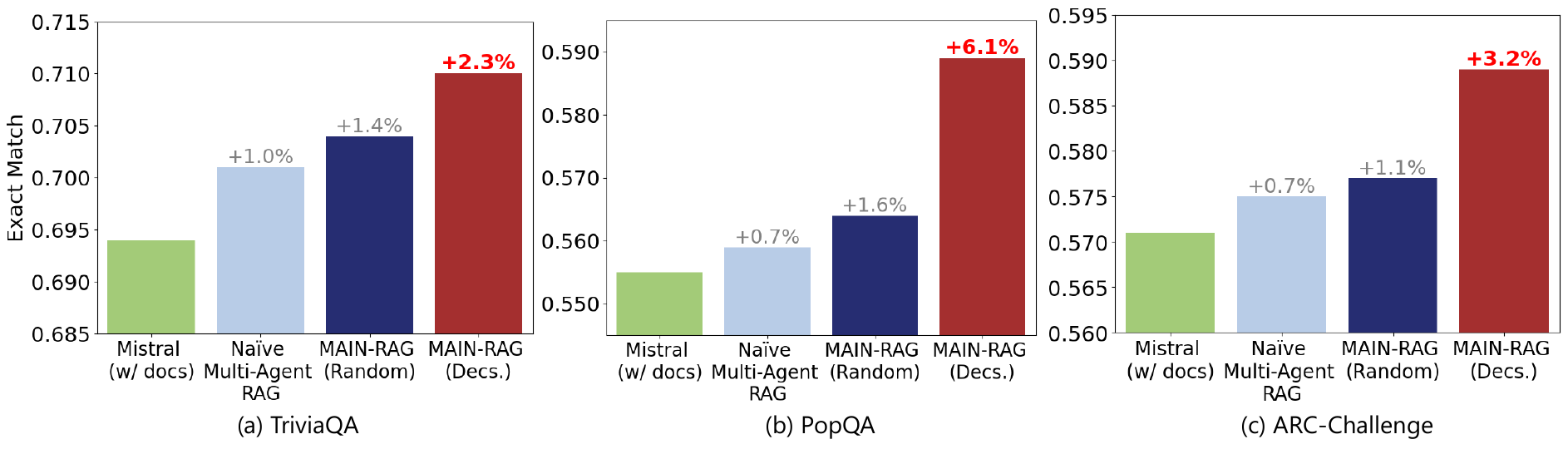}}
\vspace{-0.2cm}
\caption{Performance comparison among \Algnameabbr{} and its variant baselines on three QA benchmarks, where all three LLM agents are pre-trained $\texttt{Mistral}_{\texttt{7B}}$. Comparison based on $\texttt{Llama3}_{\texttt{8B}}$ agents is illustrated in Appendix~\ref{apx:perform_variant}.}
\label{fig:exp_performance_compa}
\vspace{-0.2cm}
\end{figure*}

\section{Experiments} 
\label{sec:exp}
In this section, we conduct experiments to evaluate the performance of \Algnameabbr{}, aiming to answer the following three research questions: \textbf{RQ1:} How does \Algnameabbr{} perform leveraging LLM agents as noisy document filter? \textbf{RQ2:} How to utilize adaptive judge bar $\tau_q$ for filtering and ranking? \textbf{RQ3:} How does $\tau_q$ influence performance?

\subsection{Tasks and Datasets}
We evaluate our \Algnameabbr{} model and various baselines across a range of downstream tasks, assessing the outputs for overall correctness. 
All evaluations are conducted in a zero-shot setting, where we provide task instructions without few-shot demonstrations~\cite{sanh2022multitask, wei2021finetuned}.

\noindent \textbf{Closed-set Task.} We evaluate \Algnameabbr{} on the ARC-Challenge dataset~\cite{clark2018think}, a multiple-choice reasoning dataset collected from scientific exams. 
We use accuracy as the evaluation metric and report results on the testing set.~\nocite{chuang2024learning, chuang2024large, wangtve, wang2024taylor, chuang2024understanding, hu2023self, hu2023st, hu2024only, wang2024assessing, yuan2024kv, jinllm, chang2020query, chang2023dispel, chang2023coda, chang2024towards}

\noindent \textbf{Open-Domain Question Answering Tasks.} 
We evaluate \Algnameabbr{} on two open-domain QA datasets: TriviaQA-unfiltered~\cite{joshi2017triviaqa} and PopQA~\cite{mallen2022not}, both of which require LLMs to answer arbitrary questions about factual knowledge. 
Since the testing set of TriviaQA-unfiltered is not publicly available, we use the validation and testing sets provided by an existing work~\cite{asaiself}, comprising 11,313 testing queries for evaluation. 
For PopQA, we utilize the long-tail subset, consisting of 1,399 rare entity queries with monthly Wikipedia page views of less than 100. 
Following prior works~\cite{mallen2022not, schick2024toolformer}, we evaluate task performance based on whether the gold answers are included in the model's generations instead of strictly requiring exact matches.

\noindent \textbf{Long-form Generation Tasks.} We conduct results on the long-form QA task ALCE-ASQA~\cite{gao2023enabling, stelmakh2022asqa} to evaluate \Algnameabbr{}. We rely on the official metrics, including correctness (str-em and rouge), and fluency measured by MAUVE (mau)~\cite{pillutla2021mauve}.

\subsection{Baselines}

\noindent \textbf{Baselines without retrievals.} We assess a range of publicly available, powerful pre-trained LLMs, including $\text{Llama2}_{\text{7B}, \text{13B}}$~\cite{touvron2023llama}, $\text{Llama3}_{\text{8B}}$~\cite{dubey2024llama}, and $\text{Mistral}_{\text{7B}}$~\cite{jiang2023mistral}, as well as instruction-tuned models like $\text{Alpaca}_{\text{7B}, \text{13B}}$~\cite{dubois2024alpacafarm}. We also compare our framework with a model trained and enhanced using private data, ${\text{Llama2-chat}_{\text{13B}}}$. Whenever possible, we use the official system prompts or instruction formats that were applied during the training process of these instruction-tuned models.

\noindent \textbf{Baselines with retrievals.} We evaluate models that incorporate retrieval, either during inference or throughout the training process.
In the first category, we include three fine-tuned models. The first is Self-RAG~\cite{asaiself}, a variant of $\text{Llama2}_{\text{7B}}$ trained to retrieve documents, generate outputs, and critically examine both retrieved passages and its own responses, expanding its vocabulary with additional reflection tokens. The second is ${\text{Llama2-FT}_{\text{7B}}}$, which is $\text{Llama2}_{\text{7B}}$ fine-tuned on the same dataset used by Self-RAG, but without the reflection tokens or retrieved passages. We also include results from a retrieval-augmented baseline, $\text{Ret-Llama2-chat}_{\text{13B}}$, which is trained on private data collected in Self-RAG and performs inference with retrieved documents. 
In the second category, we consider standard RAG baselines that do not require additional training. These methods simply prepend the top retrieved documents to the query before passing them to a pre-trained LLM (e.g., $\text{Llama2}_{\text{7B}, \text{13B}}$, $\text{Alpaca}_{\text{7B}, \text{13B}}$, $\text{Llama3}_{\text{8B}}$, $\text{Mistral}_{\text{7B}}$), using the same retriever as in our system. We also consider two variants of \Algnameabbr{}: \textbf{Naïve Multi-agent RAG}: This \Algnameabbr{} variant replaces \texttt{Agent-2}'s role with a simple natural language judgment of "Yes" or "No"; \textbf{\Algnameabbr{} (Random)}: In this variant, after scoring and filtering, the orders of remaining documents are randomized.

\subsection{Experimental Settings}

As a training-free RAG framework, the three agents in \Algnameabbr{} can be instantiated by different pre-trained LLMs. 
As default settings, we instantiate all three agents by pre-trained $\text{Mistral}_{\text{7B}}$~\cite{jiang2023mistral} and $\text{Llama3}_{\text{8B}}$ without further tuning. We employ the pre-trained Contriever-MS MARCO~\cite{izacard2021unsupervised} as the default retriever model, retrieving up to twenty documents from each query for \Algnameabbr{} to filter. We use greedy generation for all our experiments.

\renewcommand{\arraystretch}{0.6} 
\begin{table*}[t]
\centering
\caption{Overall experimental results on four tasks. \textbf{Bold} numbers refer to the best performance among baselines without retrieval and training-free baselines, and \underline{underline} numbers refer to the second-best performance. \textcolor{gray}{\textbf{Gray bold}} numbers refer to the best performance among proprietary models and training-based baselines. $^{*}$ indicates concurrent results conducted by recent works or original papers. For the metrics, \textit{acc}, \textit{em}, \textit{rg}, and \textit{mau} denote \textit{accuracy}, \textit{str-em}, \textit{rouge}, and \textit{MAUVE}, respectively.}
\begin{tabularx}{\textwidth}
{%
    p{3.8cm}  
    >{\centering\arraybackslash}p{2cm}  
    >{\centering\arraybackslash}p{2cm}  
    >{\centering\arraybackslash}p{2cm}  
    >{\centering\arraybackslash}p{0.8cm}  
    >{\centering\arraybackslash}p{0.8cm}  
    >{\centering\arraybackslash}p{0.8cm}  
}
    \toprule
    & TriviaQA & PopQA & ARC-C & \multicolumn{3}{c}{ASQA} \\
    & (acc) & (acc) & (acc) & (em) & (rg) & (mau) \\
    \midrule
    \multicolumn{7}{c}{\textit{LMs with proprietary data}} \\
    ${\text{Llama2-chat}_{\text{13B}}}^{*}$ & 59.3 & 20.0 & 38.4 & 22.4 & 29.6 & 28.6 \\
    ${\text{Ret-Llama2-chat}_{\text{13B}}}^{*}$ & 59.8 & 51.8 & 37.9 & \textbf{\textcolor{gray}{32.8}} & 34.8 & 43.8 \\
    \midrule
    \multicolumn{7}{c}{\textit{Baselines with retrieval (training-based)}} \\
    ${\text{Llama2-FT}_{\text{7B}}}^{*}$ & 57.3 & 48.7 & 65.8 & 31.0 & \textbf{\textcolor{gray}{35.8}} & 51.2 \\
    ${\text{Self-RAG}_{\text{7B}}}^{*}$ & \textbf{\textcolor{gray}{66.4}} & \textbf{\textcolor{gray}{54.9}} & \textbf{\textcolor{gray}{67.3}} & 30.0 & 35.7 & \textbf{\textcolor{gray}{74.3}} \\
    \midrule
    \multicolumn{7}{c}{\textit{Baselines without retrieval}} \\
    ${\text{Llama2}_{\text{7B}}}^{*}$ & 30.5 & 14.7 & 21.8 & 7.9 & 15.3 & 19.0 \\
    ${\text{Alpaca}_{\text{7B}}}^{*}$ & 54.5 & 23.6 & 45.0 & 18.8 & 29.4 & 61.7 \\
    ${\text{Llama2}_{\text{13B}}}^{*}$ & 38.5 & 14.7 & 29.4 & 7.2 & 12.4 & 16.0 \\
    ${\text{Alpaca}_{\text{13B}}}^{*}$ & 61.3 & 24.4 & 54.9 & 22.9 & 32.0 & \textbf{70.6} \\
    $\text{Mistral}_{\text{7B}}$ & 54.8 & 26.2 & 55.5 & 11.2  & 18.1 & 27.6 \\
    $\text{Llama3}_{\text{8B}}$ & 68.4 & 29.2 & 58.8 & 19.4 & 30.3 & 54.3 \\
    \midrule
    \multicolumn{7}{c}{\textit{Baselines with retrieval (training-free)}} \\
    $\text{Llama2}_{\text{7B}}$ & 68.9 & 50.9 & 51.0 & 16.2 & 23.4 & 33.1 \\
    ${\text{Alpaca}_{\text{7B}}}^{*}$ & 64.1 & 46.7 & 48.0 & 30.9 & 33.3 & 57.9 \\
    ${\text{Llama2}_{\text{13B}}}^{*}$ & 47.0 & 45.7 & 26.0 & 16.3 & 20.5 & 24.7 \\
    ${\text{Alpaca}_{\text{13B}}}^{*}$ & 66.9 & 46.1 & 57.6 & 34.8 & \underline{36.7} & 56.6 \\
    $\text{Mistral}_{\text{7B}}$ & 69.4 & 55.5 & 57.1 & 32.4 & 34.8 & 54.3 \\
    $\text{Llama3}_{\text{8B}}$ & \underline{73.1} & \underline{61.8} & 55.6 & \underline{37.1} & 36.5 & \underline{63.0} \\
    \noalign{\vskip 0.05cm}
    \hdashline
    \noalign{\vskip 0.05cm}
    \Algnameabbr{}-$\text{Mistral}_{\text{7B}}$ & 71.0 & 58.9 & \underline{58.9} & 35.7 & 36.2 & 60.0 \\
    \Algnameabbr{}-$\text{Llama3}_{\text{8B}}$ & \textbf{74.1} & \textbf{64.0} & \textbf{61.9} & \textbf{39.2} & \textbf{42.0} & \textbf{70.6} \\
    \bottomrule
\end{tabularx}%
\label{tab:exp_results}
\end{table*}

\subsection{Quantitative Analysis (RQ1)}
\label{sec:quantitative}
We evaluated the performance of our proposed \Algnameabbr{} framework and baselines across four well-known QA benchmarks, where \Algnameabbr{} (Decs.) refers to our method that orders documents in descending order after scoring and filtering, as illustrated in Figure~\ref{fig:exp_performance_compa}, Table~\ref{tab:exp_results}, and Appendix~\ref{apx:perform_variant}. 
Our results demonstrate that \Algnameabbr{} outperforms all training-free and without retrieval baselines by margins up to $\texttt{6.1}\%$ (with $\texttt{Mistral}_{\texttt{7B}}$) and $\texttt{12.0}\%$ (with $\texttt{Llama3}_{\texttt{8B}}$) in all four benchmarks, as shown in Table~\ref{tab:exp_results}. Notably, the questions in PopQA heavily rely on external knowledge to enable pre-trained LLMs to generate accurate answers. 
In this case, \Algnameabbr{} exhibits a significant advantage over the baselines, because the retriever is not fine-tuned on the target question sets and may retrieve a large number of noisy candidate documents.
Compared with training-based baselines, our training-free \Algnameabbr{} framework can bridge the performance gap in TriviaQA and PopQA datasets. We also found that on the metrics for rough (rg), \Algnameabbr{}-$\text{Mistral}_{\text{7B}}$ occasionally outperforms the two training-based baselines, ${\text{Self-RAG}_{\text{7B}}}$ and ${\text{Llama2-FT}_{\text{7B}}}$, showing the potential of improving pre-trained LLMs to outperform resource-consuming fine-tuning methods.

\subsection{Ablation Studies on Adaptive Judge Bar $\tau_q$ for Filtering and Ranking (RQ2)}

We assess the effectiveness of the adaptive judgment bar $\tau_q$ by comparing the default $\tau_q$ with variations adjusted by different scales of standard deviation, $\tau_q - n \cdot \sigma$. As mentioned in Section~\ref{sec:adaptive_bar}, the purpose of these adjustments is to relax the filtering threshold when the recall rate of relevant documents is low, potentially preventing the omission of critical external information required to support LLMs in question answering. Despite its flexibility, our experiments demonstrate that the default $\tau_q$ generally performs well in filtering noisy documents. As shown in Table~\ref{tab:judge_bar_comp}, while the adjusted variants randomly achieve better performance, the default $\tau_q$ consistently ranks at least second-best across three benchmarks and two different pre-trained LLMs, indicating its practicality.

After filtering out irrelevant or noisy documents, the remaining candidate documents can be sorted in either descending or ascending order. As shown in Table~\ref{tab:judge_bar_comp}, \Algnameabbr{} defaults to descending order, consistently achieving better performance compared to ascending order. This result aligns with findings from prior work, which suggests that LLMs tend to prioritize information presented at the beginning of the input~\cite{liu2024lost}.

\renewcommand{\arraystretch}{1.0} 
\begin{table}[tb]
\centering
\small
\caption{Ablation studies of $\tau_q$ and document ordering. \textbf{Bold} numbers indicate the best result, and \underline{underline} numbers indicate the second-best result.}
\begin{tabular}{l c c c}
    \toprule
    & TriviaQA & PopQA & ARC-C  \\
    & (acc) & (acc)  & (acc) \\
    \midrule
    \multicolumn{4}{c}{$\text{Mistral}_{\text{7B}}$} \\
    \Algnameabbr{} (Decs.) & \underline{71.0} & \textbf{58.9} & \underline{58.9}  \\
    \noalign{\vskip 0.05cm}
    \hdashline
    \noalign{\vskip 0.05cm}
    \Algnameabbr{} (Asc.) & 70.2 & 53.5 & 57.4 \\ 
    \Algnameabbr{} ($\tau_q - 0.5 \cdot \sigma)$ & \textbf{71.2} & \underline{58.6} & \textbf{59.0} \\
    \Algnameabbr{} ($\tau_q - 1.0 \cdot \sigma)$ & 70.8 & 58.0 & 58.5 \\
    \Algnameabbr{} ($\tau_q - 1.5 \cdot \sigma)$ & 70.4 & 58.4 & 57.7 \\
    \midrule
    \multicolumn{4}{c}{$\text{Llama3}_{\text{8B}}$} \\
    \Algnameabbr{} (Decs.) & \underline{74.1} & \textbf{64.0} & \textbf{61.9} \\ 
    \noalign{\vskip 0.05cm}
    \hdashline
    \noalign{\vskip 0.05cm}
    \Algnameabbr{} (Asc.) & 73.6 & \underline{63.5} & \underline{60.7} \\ 
    \Algnameabbr{} ($\tau_q - 0.5 \cdot \sigma)$ & \underline{74.1} & \textbf{64.0} & 58.6 \\
    \Algnameabbr{} ($\tau_q - 1.0 \cdot \sigma)$ & \underline{74.1} & 63.3 & 58.9 \\
    \Algnameabbr{} ($\tau_q - 1.5 \cdot \sigma)$ & \textbf{74.3} & \textbf{64.0} & 57.2 \\
    \bottomrule
\end{tabular}%
\vspace{-0.2cm}
\label{tab:judge_bar_comp}
\end{table}

\begin{figure*}[ht]
\centering
\small
\fcolorbox{black}{gray!10}{\parbox{.95\linewidth}
    {
        \hspace{-1cm}\textbf{Case Study 1} \\
        \textbf{Question:} In what city was Montxu Miranda born? \\
        \textbf{Adaptive Judge Bar $\tau_q$}: \textbf{\texttt{9.575}} \\
        \underline{Filtered and Ordered Documents}: Montxu Miranda Montxu Miranda Díez (born 27 December 1976 in Santurce) is a Spanish pole vaulter. His personal best of 5.81 metres, achieved in September 2000 in Barcelona, is still the standing Spanish national record. ... He studied at the Colegio San Calixto, then later pursued a career in Political Sciences at the Higher University of San Andrés in La Paz. \\
        \underline{Ground Truth:} "Santurtzi", "Santurce" \\
        \underline{\textbf{LLM Answer:}} Montxu Miranda was born in Santurce. \textbf{(\textcolor{darkgreen}{correct})} \\
        \hangindent=1cm \hangafter=0
        
        \hspace{-1cm}\textbf{Case Study 2} \\
        \textbf{Question:} What is the capital of Gmina Czorsztyn? \\
        \textbf{Adaptive Judge Bar $\tau_q$}: \textbf{\texttt{-8.425}} \\
        \underline{Filtered and Ordered Documents}: Gmina Wolsztyn is an urban-rural gmina (administrative district) in Wolsztyn County, ... Sromowce Wyżne is a village in the administrative district of Gmina Czorsztyn, within Nowy Targ County, Lesser Poland Voivodeship, in southern Poland, close to the border with Slovakia. It lies approximately 8 km south-east of Maniowy, 25 km east of Nowy Targ, ... \\
        \underline{Ground Truth:} "Maniowy" \\
        \underline{\textbf{LLM Answer:}} The capital of Gmina Czorsztyn is Maniowy. \textbf{(\textcolor{darkgreen}{correct})} \\
        \hangindent=1cm \hangafter=0

        \hspace{-1cm}\textbf{Case Study 3} \\
        \textbf{Question:} What is Arcangelo Ghisleri's occupation? \\
        \textbf{Adaptive Judge Bar $\tau_q$}: \textbf{\texttt{0.4875}} \\
        \underline{Filtered and Ordered Documents}: S. Michele Arcangelo, archangel in Jewish, Christian, and Islamic teachings ; Andrea di Cione Arcangelo (1308–1368), Italian painter, sculptor, and architect active in Florence ; Antonio di Arcangelo, Italian painter, active in Florence in a Renaissance style, between 1520 and 1538 ; Arcangelo Califano (1730–1750), baroque composer and cellist... \\
        \underline{Ground Truth:} "journalist", "journo", "journalists" \\
        \underline{\textbf{LLM Answer:}} Arcangelo Ghisleri was an Italian geographer, writer, and Socialist politician. \textbf{(\textcolor{darkred}{wrong})} \\
        \hangindent=1cm \hangafter=0
    }
}
\caption{Case Study: \textbf{Adaptive Judge Bar $\tau_q$} (Dataset: \textbf{PopQA}; \texttt{LLM Agents}: \textbf{$\texttt{Mistral}_{\texttt{7B}}$})}
\label{fig:case_popqa}
\end{figure*}

\subsection{Case Studies of Different Adaptive Judge Bar $\tau_q$ (RQ3)}

\Algnameabbr{} involves adaptive judge bar $\tau_q$ to approximate optimal judge bars of each query by averaging relevant scores over retrieved documents for a query. This approach is inspired by our observation of distinct score distributions between the most relevant document set and the least relevant document set, as discussed in Section~\ref{sec:adaptive_bar}. 
From Figure~\ref{fig:score_dist}, we observe that \texttt{Agent-2} assigns confidently high relevance scores to related documents, resulting in a skewed-high score distribution. In contrast, while \texttt{Agent-2} scores noisy documents with a more uniform distribution, the lowest scores for noisy documents are significantly lower than those for related documents. This disparity allows the filtering mechanism to improve the prediction accuracy of \texttt{Agent-3}, regardless of whether $\tau_q$ is set relatively high or low. The correlation between $\tau_q$ and performance can be observed in Figure~\ref{fig:case_popqa} and further discussed in Appendix~\ref{apx:case}.

\section{Conclusion and Future Work}
In this work, we address the challenges of noisy document retrieval in RAG by introducing a training-free, multi-agent framework, \Algnameabbr{}. 
Our approach employs multiple LLM agents to collaboratively filter and rank retrieved documents, enhancing the recall of relevant information while minimizing irrelevant content. 
Specifically, \Algnameabbr{} utilizes an adaptive judge bar that dynamically adjusts based on the score distribution of relevant and noisy documents in different queries. 
Experimental results demonstrate that \Algnameabbr{} consistently outperforms training-free RAG baselines across various QA benchmarks.
Regarding future directions, the \Algnameabbr{} framework unveils several potential facets that merit further exploration, such as integrating with a more fine-grained adaptive judge bar, extending the approach to other tasks beyond question answering, and incorporating human feedback or tuning-based approaches to enhance the efficacy of document filtering.

\clearpage
\section{Limitations}

We conduct experiments on four datasets using two different pre-trained LLM architectures. These experiments primarily focus on LLM inference with retrieved external documents. We did not consider other research questions in RAG or prompting language models such as context compression~\cite{xu2024recomp,li2024prompt}, documents position and ordering~\cite{liu2024lost,xu-etal-2024-context}, decoding methods~\cite{li-etal-2023-contrastive,shi-etal-2024-trusting,xu2023context}. The choice of retrievers and rerankers is also orthogonal to our main research agenda~\cite[\emph{inter alia}]{lin2022pretrained,asai2024openscholar,xu2024rankmamba,xu2024state}. 

We acknowledge that LLM inference under RAG workflow contributes to carbon emissions, representing a potential limitation and environmental risk of our work. To mitigate this, we aim to reduce the need for repetitive experiments by ensuring more predictable outcomes and implementing controlled experimental settings.

\bibliography{arxiv}
\bibliographystyle{arxiv}

\clearpage
\appendix

\onecolumn
\section*{Appendix}
\section{Computation Infrastructure}
\label{apx:infra}
For a fair comparison of evaluation, the experiments are conducted based on the following physical computing infrastructure in Table~\ref{tab:computing_infrastructure}.

\begin{table}[h!]
\centering
\caption{Computing infrastructure for the experiments.}
\label{tab:computing_infrastructure}
    \begin{tabular}{l c}
        \toprule
        Device Attribute & Spec \\
        \midrule
        Computing Infrastructure & GPU \\
        GPU Model & Nvidia-A100 \\
        GPU Number & 4 \\
        GPU Memory & 80 GB \\
        \bottomrule
    \end{tabular}
\end{table}

\section{Performance Comparison among \Algnameabbr{} and Its Variant Baselines}
\label{apx:perform_variant}

\begin{figure*}[hb]
\centerline{\includegraphics[width=1.\textwidth]{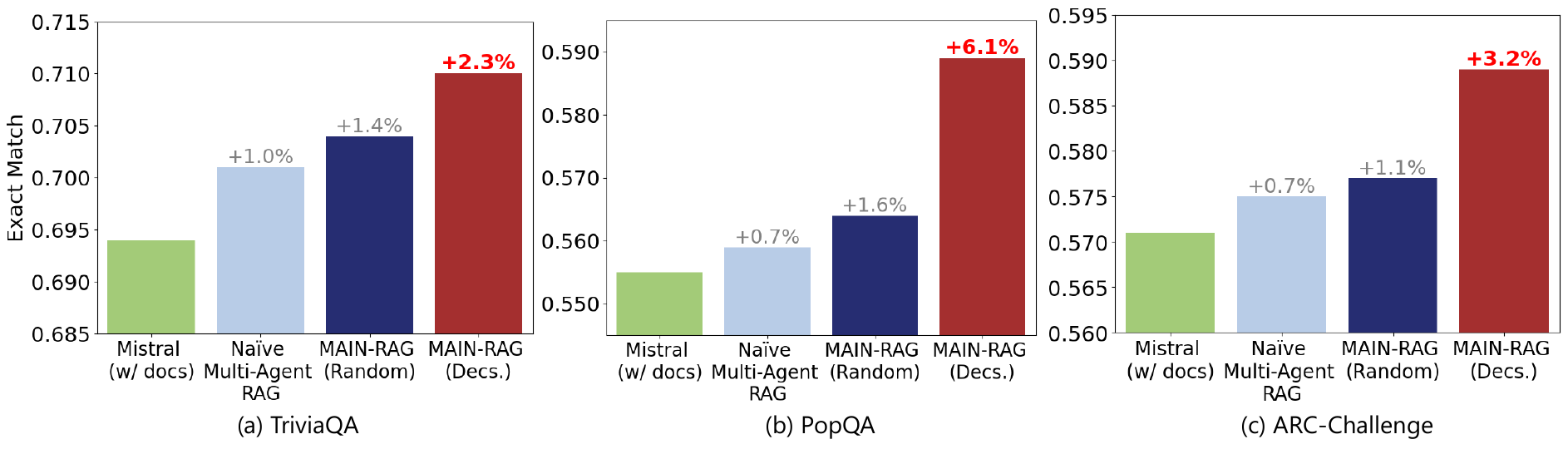}}
\caption{Performance comparison among \Algnameabbr{} and its variant baselines on three QA benchmarks, where all three LLM agents are pre-trained $\texttt{Mistral}_{\texttt{7B}}$.}
\label{apx:fig:exp_performance_compa}
\end{figure*}

\begin{figure*}[hb]
\centerline{\includegraphics[width=1.\textwidth]{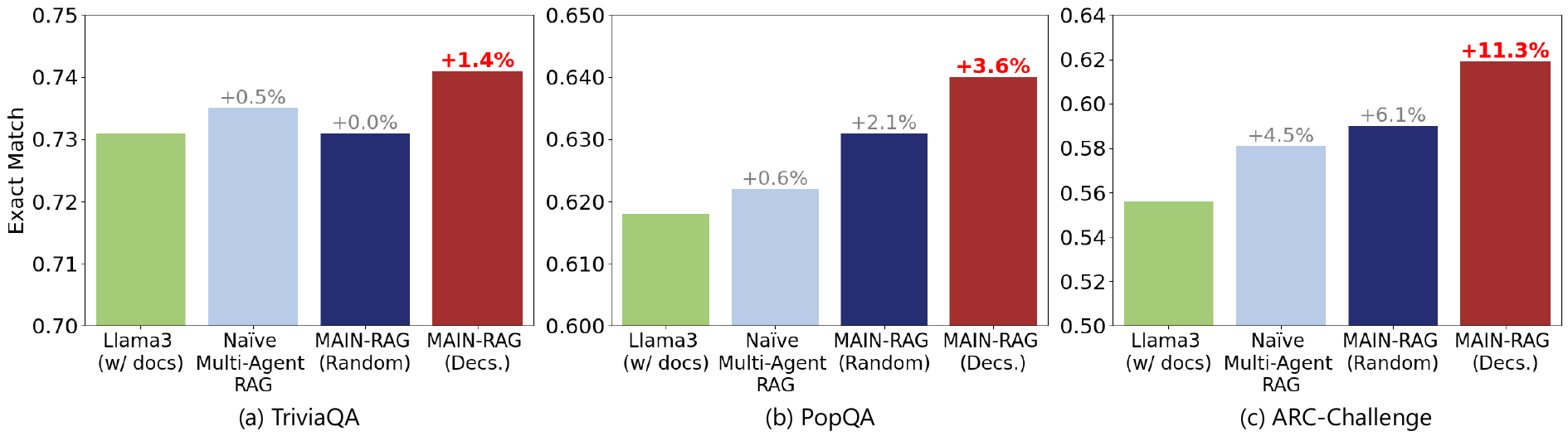}}
\caption{Performance comparison among \Algnameabbr{} and its variant baselines on three QA benchmarks, where all three LLM agents are pre-trained $\texttt{Llama3}_{\texttt{8B}}$.}
\label{apx:fig:exp_performance_compa_llama}
\end{figure*}

Our results demonstrate that \Algnameabbr{} outperforms all training-free, without retrieval, and \Algnameabbr{} variant baselines by margins up to $\texttt{6.1}\%$ (with $\texttt{Mistral}_{\texttt{7B}}$) and $\texttt{12.0}\%$ (with $\texttt{Llama3}_{\texttt{8B}}$) in all four benchmarks, as shown in Table~\ref{tab:exp_results}, Figure~\ref{apx:fig:exp_performance_compa}, and Figure~\ref{apx:fig:exp_performance_compa_llama}. Notably, the questions in PopQA heavily rely on external knowledge to enable pre-trained LLMs to generate accurate answers. 
In this case, \Algnameabbr{} exhibits a significant advantage over the baselines, because the retriever is not fine-tuned on the target question sets and may retrieve a large number of noisy candidate documents.

\section{System Instructions of \texttt{Agent-1 (Predictor)}, \texttt{Agent-2 (Judge)}, and \texttt{Agent-3 (Final-Predictor)}}
\label{apx:agent_prompt}

We present the system instructions of \texttt{Agent-1 (Predictor)}, \texttt{Agent-2 (Judge)}, and \texttt{Agent-3 (Final-Predictor)} in Figure~\ref{apx:fig:agent-prompt} in \Algnameabbr{} framework.

\begin{figure*}[bh]
\fcolorbox{black}{lightgray!15}{\parbox{.95\linewidth}
{
\textbf{System Instruction (\texttt{Agent-1 (Predictor)}):} \\
You are an accurate and reliable AI assistant that can answer questions with the help of external documents. You should only provide the correct answer without repeating the question and instruction.
\\
\vspace{1mm}
\!\!\!\!\!\! \hrulefill
\\
\textbf{System Instruction (\texttt{Agent-2 (Judge)}):} \\
You are a noisy document evaluator that can judge if the external document is noisy for the query with unrelated or misleading information. Given a retrieved Document, a Question, and an Answer generated by an LLM (LLM Answer), you should judge whether both the following two conditions are reached: (1) the Document provides specific information for answering the Question; (2) the LLM Answer directly answers the question based on the retrieved Document. Please note that external documents may contain noisy or factually incorrect information. If the information in the document does not contain the answer, you should point it out with evidence. You should answer with "Yes" or "No" with evidence of your judgment, where "No" means one of the conditions (1) and (2) are unreached and indicates it is a noisy document.
\\
\vspace{1mm}
\!\!\!\!\!\! \hrulefill
\\
\textbf{System Instruction (\texttt{Agent-3 (Final-Predictor)}):} \\
You are an accurate and reliable AI assistant that can answer questions with the help of external documents. You should only provide the correct answer without repeating the question and instruction.
\vspace{1mm}
}}
\caption{\small System instructions of \textbf{\texttt{Agent-1 (Predictor)}}, \textbf{\texttt{Agent-2 (Judge)}}, and \textbf{\texttt{Agent-3 (Final-Predictor)}}.}
\label{apx:fig:agent-prompt}
\end{figure*}

\section{Case Studies of Different Adaptive Judge Bar $\tau_q$ in \Algnameabbr{}}
\label{apx:case}
We demonstrate the case studies of the documents filtered and ranked by \Algnameabbr{} under different adaptive judge bars $\tau_q$. The demonstrations show that \texttt{Agent-2 (Judge)}, pre-trained LLMs, is able to provide convincing judgment ability for the queries with high adaptive judge bar $\tau_q$ (confident), but suffers in those low adaptive judge bar queries (unconfident). The case studies on \texttt{PopQA} and \texttt{TriviaQA} are illustrated in Figure~\ref{apx:fig:case_popqa_mistral}-\ref{apx:fig:case_trivia_llama3}.

\begin{figure*}[ht]
\centering
\fcolorbox{black}{gray!10}{\parbox{.95\linewidth}
    {
        \hspace{-1cm}\textbf{Case Study 1} \\
        \textbf{Question:} In what city was Montxu Miranda born? \\
        \textbf{Adaptive Judge Bar $\tau_q$}: \textbf{\texttt{9.575}} \\
        \underline{Filtered and Ordered Documents}: Montxu Miranda Montxu Miranda Díez (born 27 December 1976 in Santurce) is a Spanish pole vaulter. His personal best of 5.81 metres, achieved in September 2000 in Barcelona, is still the standing Spanish national record. Miranda was born in 1981 in Buenaventura, Colombia and was abandoned by his mother when he was one month old. At age 9, a precocious Miranda began a quest to locate his mother. He hitched rides on truck beds over the course of hundreds of miles and was able to find his uncle working at a construction site. His uncle informed him that if he was really the little boy his sister gave away, then he should have a birthmark on his leg whereas Miranda unveiled the 2-inch long circular proof. The man led him to his mother's new home, only to be abandoned again by his mother. ... \\
        \underline{Ground Truth:} "Santurtzi", "Santurce" \\
        \underline{\textbf{LLM Answer:}} Montxu Miranda was born in Santurce. \textbf{(\textcolor{darkgreen}{correct})} \\
        \hangindent=1cm \hangafter=0
        
        \hspace{-1cm}\textbf{Case Study 2} \\
        \textbf{Question:} What is the capital of Gmina Czorsztyn? \\
        \textbf{Adaptive Judge Bar $\tau_q$}: \textbf{\texttt{-8.425}} \\
        \underline{Filtered and Ordered Documents}: Gmina Wolsztyn is an urban-rural gmina (administrative district) in Wolsztyn County, ... Sromowce Wyżne is a village in the administrative district of Gmina Czorsztyn, within Nowy Targ County, Lesser Poland Voivodeship, in southern Poland, close to the border with Slovakia. It lies approximately 8 km south-east of Maniowy, 25 km east of Nowy Targ, and 78 km south-east of the regional capital Kraków. The village has a population of 1,100. Czosnów is a village in Nowy Dwór Mazowiecki County, Masovian Voivodeship, in east-central Poland. It is the seat of the gmina (administrative district) called Gmina Czosnów. It lies approximately 7 km southeast of Nowy Dwór Mazowiecki and 26 km northwest of Warsaw. The village has a population of 420. ... \\
        \underline{Ground Truth:} "Maniowy" \\
        \underline{\textbf{LLM Answer:}} The capital of Gmina Czorsztyn is Maniowy. \textbf{(\textcolor{darkgreen}{correct})} \\
        \hangindent=1cm \hangafter=0

        \hspace{-1cm}\textbf{Case Study 3} \\
        \textbf{Question:} What is Arcangelo Ghisleri's occupation? \\
        \textbf{Adaptive Judge Bar $\tau_q$}: \textbf{\texttt{0.4875}} \\
        \underline{Filtered and Ordered Documents}: S. Michele Arcangelo, archangel in Jewish, Christian, and Islamic teachings ; Andrea di Cione Arcangelo (1308–1368), Italian painter, sculptor, and architect active in Florence ; Antonio di Arcangelo, Italian painter, active in Florence in a Renaissance style, between 1520 and 1538 ; Arcangelo Califano (1730–1750), baroque composer and cellist ; Arcangelo Placenza da Calatafimi, (1390–1460) venerated Italian Franciscan friar and preacher ; Arcangelo Canetoli (1460–1513), venerated Catholic priest ; Arcangelo Cascieri (1902–1997), influential sculptor, major figure in Boston Architectural College in Boston, Massachusetts ; Arcangelo di Cola (active 1416-1429) Italian late-Gothic painter ; ... \\
        \underline{Ground Truth:} "journalist", "journo", "journalists" \\
        \underline{\textbf{LLM Answer:}} Arcangelo Ghisleri was an Italian geographer, writer, and Socialist politician. \textbf{(\textcolor{darkred}{wrong})} \\
        \hangindent=1cm \hangafter=0
    }
}
\caption{Case Study: \textbf{Adaptive Judge Bar $\tau_q$} (Dataset: \textbf{PopQA}; \texttt{LLM Agents}: \textbf{$\texttt{Mistral}_{\texttt{7B}}$})}
\label{apx:fig:case_popqa_mistral}
\end{figure*}

\begin{figure*}[ht]
\centering
\fcolorbox{black}{gray!10}{\parbox{.95\linewidth}
    {
        \hspace{-1cm}\textbf{Case Study 1} \\
        \textbf{Question:} In what city was Fernando García born? \\
        \textbf{Adaptive Judge Bar $\tau_q$}: \textbf{\texttt{6.234}} \\
        \underline{Filtered and Ordered Documents}: Fernando García (born July 4, 1930 in Santiago, Chile) is a Chilean composer. Active since 1956 he has done orchestral music, chamber music, etc. He studied with Juan Orrego-Salas and Gustavo Becerra-Schmidt, among others. His style is strongly influenced by serialism and aleatoric procedures. He also played a role in the beginnings of electroacoustic music in Chile, after a trip he made to France in the early 1950s where he heard musique concrete. He worked for the Instituto de Extensión Musical of the University of Chile, and in 1962 he premiered his most important piece, the cantata América Insurrecta, which won an award at the Chilean Music Festival. After the Chilean coup-d'état, he was forced into exile, first in Perú (1973-1979), and then in Cuba (1979-1990). ... \\
        \underline{Ground Truth:} "Santiago", "Santiago de Chile", "Santiago, Chile" \\
        \underline{\textbf{LLM Answer:}} According to the document, Fernando García was born in Santiago, Chile. \textbf{(\textcolor{darkgreen}{correct})} \\
        \hangindent=1cm \hangafter=0
        
        \hspace{-1cm}\textbf{Case Study 2} \\
        \textbf{Question:} In what country is Ittamalliyagoda? \\
        \textbf{Adaptive Judge Bar $\tau_q$}: \textbf{\texttt{-6.88}} \\
        \underline{Filtered and Ordered Documents}: Ittamalliyagoda Ittamalliyagoda is a village in Sri Lanka. It is located within Central Province. Loluwagoda Loluwagoda is a well known village in Sri Lanka which is situated on Nittambuwa - Giriulla 221 main road between Mirigama and Giriulla towns. Main religion of the people in the village is Theravada Buddhism.The old Ramanna Nikaya temple in the village (Sri Mahendraramaya) was well-known among Buddhist monks in the area as a hidden and quiet temple which is not promoting social services or festivals in order to obtain donations. A group from the villagers annually donate medicine to the Mitirigala Nissarana Vanaya with a group from Atambe village. Like the most villages in Gampaha district, the population Loluwagoda is a village in the Gampaha District, Sri Lanka. It is located on the Nittambuwa - Giriulla 221 main road, between the towns of Mirigama and Giriulla. ... \\
        \underline{Ground Truth:} "Sri Lanka", "Democratic Socialist Republic of Sri Lanka" \\
        \underline{\textbf{LLM Answer:}} According to the document, Ittamalliyagoda is a village in Sri Lanka. \textbf{(\textcolor{darkgreen}{correct})} \\
        \hangindent=1cm \hangafter=0

        \hspace{-1cm}\textbf{Case Study 3} \\
        \textbf{Question:} What is Andreas Rüdiger's occupation? \\
        \textbf{Adaptive Judge Bar $\tau_q$}: \textbf{\texttt{0.40625}} \\
        \underline{Filtered and Ordered Documents}: Johannes Andreas Rüdiger (1 November 1673 – 6 June 1731) was a German philosopher and physicist. Antonio Rüdiger (born 3 March 1993) is a German professional footballer who plays as a centre-back for club Chelsea and the Germany national team. Rüdiger began his career at VfB Stuttgart, representing their reserves in 3. Liga and the first team in the Bundesliga. In 2015 he joined Roma, initially on loan and a year later for a €9 million fee. He was signed by Chelsea in 2017 for an estimated £27 million, where he won the FA Cup in his first season, followed by the UEFA Europa League in his second season, and the UEFA Champions League in 2021. He made his international debut for Germany in March 2014. ... \\
        \underline{Ground Truth:} "philosopher" \\
        \underline{\textbf{LLM Answer:}} According to the document, Andreas Rüdiger (born 1993) is a German professional footballer who plays as a centre-back for club Chelsea and the Germany national team. \textbf{(\textcolor{darkred}{wrong})} \\
        \hangindent=1cm \hangafter=0
    }
}
\caption{Case Study: \textbf{Adaptive Judge Bar $\tau_q$} (Dataset: \textbf{PopQA}; \texttt{LLM Agents}: \textbf{$\texttt{Llama3}_{\texttt{8B}}$})}
\label{apx:fig:case_popqa_llama3}
\end{figure*}

\begin{figure*}[ht]
\centering
\fcolorbox{black}{gray!10}{\parbox{.95\linewidth}
    {
        \hspace{-1cm}\textbf{Case Study 1} \\
        \textbf{Question:} In which US state is New Haven? \\
        \textbf{Adaptive Judge Bar $\tau_q$}: \textbf{\texttt{16.13}} \\
        \underline{Filtered and Ordered Documents}: New Haven, Connecticut New Haven is a coastal city in the U.S. state of Connecticut. It is located on New Haven Harbor on the northern shore of Long Island Sound in New Haven County, Connecticut, and is part of the New York metropolitan area. With a population of 129,779 as determined by the 2010 United States Census, it is the second-largest city in Connecticut after Bridgeport. New Haven is the principal municipality of Greater New Haven, which had a total population of 862,477 in 2010. New Haven was the first planned city in America. Founded in 1638 by English Puritans, New Haven, Vermont New Haven is a town in Addison County, Vermont, United States. The population was 1,727 at the 2010 census. ... \\
        \underline{Ground Truth:} "richest places in connecticut", "Connecticut", "Connectecut" \\
        \underline{\textbf{LLM Answer:}} New Haven is located in the US state of Connecticut. \textbf{(\textcolor{darkgreen}{correct})} \\
        \hangindent=1cm \hangafter=0
        
        \hspace{-1cm}\textbf{Case Study 2} \\
        \textbf{Question:} Where is the area of 127,000 square kilometres (49,000 sq miles) in central South Australia, where public are not admitted under Australian Defence Force regulations? \\
        \textbf{Adaptive Judge Bar $\tau_q$}: \textbf{\texttt{-10.675}} \\
        \underline{Filtered and Ordered Documents}: would double the area of land in South Australia under environmental protection to 1.8 million hectares. The area includes 390 species of plants and a large number of habitats for rare species of animals and birds. In May 2011, after years of negotiations, Rann joined with Federal Defence Minister Stephen Smith and Resources Minister Martin Ferguson to announce that large areas of the Woomera Prohibited Area, the largest defence testing reserve in the world, would be opened up for mining, allowing the future exploitation of mineral deposits estimated at billions of dollars. ... \\
        \underline{Ground Truth:} "woomera disambiguation", "Woomera" \\
        \underline{\textbf{LLM Answer:}} The area of 127,000 square kilometres (49,000 sq miles) in central South Australia, where public are not admitted under Australian Defence Force regulations, is the Woomera Prohibited Area. \textbf{(\textcolor{darkgreen}{correct})} \\
        \hangindent=1cm \hangafter=0

        \hspace{-1cm}\textbf{Case Study 3} \\
        \textbf{Question:} What name is given to an alcoholic drink that is taken in an effort to cure a hangover? \\
        \textbf{Adaptive Judge Bar $\tau_q$}: \textbf{\texttt{0.49375}} \\
        \underline{Filtered and Ordered Documents}: Hangover drink in Korea Hangover drinks are drinks sold in South Korea that are consumed to ease the hangover after heavy drinking. Hangover drinks are consumed before a heavy drinking session and an ingredient in them is said to break down a toxin produced in our liver when drinking and also reduce the effect of alcohol on our brain's neurotransmitters. Get-together among co-workers or students are very common in Korea. A lot of people in Korea consume the hangover drinks before they attend the get-together so they can get out of the hangover quickly on the next day. hangover cure. ... \\
        \underline{Ground Truth:} "hair of dog", "alcohol readministration", "recovery drinking", "hair off dogs back" \\
        \underline{\textbf{LLM Answer:}} The name given to an alcoholic drink that is taken in an effort to cure a hangover is "hangover drink" or "hangover cure." In Korea, it is called "haejangsul." \textbf{(\textcolor{darkred}{wrong})} \\
        \hangindent=1cm \hangafter=0
    }
}
\caption{Case Study: \textbf{Adaptive Judge Bar $\tau_q$} (Dataset: \textbf{TriviaQA}; \texttt{LLM Agents}: \textbf{$\texttt{Mistral}_{\texttt{7B}}$})}
\label{apx:fig:case_trivia_mistral}
\end{figure*}

\begin{figure*}[ht]
\centering
\fcolorbox{black}{gray!10}{\parbox{.95\linewidth}
    {
        \hspace{-1cm}\textbf{Case Study 1} \\
        \textbf{Question:} In Ancient Greece what was a hoplite? \\
        \textbf{Adaptive Judge Bar $\tau_q$}: \textbf{\texttt{8.522}} \\
        \underline{Filtered and Ordered Documents}: Hoplite formation in art The hoplites were soldiers from Ancient Greece who were usually free citizens. They had a very uniform and distinct appearance; specifically they were armed with a spear ("dory") in their right hand and a heavy round shield in their left. Hoplite soldiers were organized in battle into the "Phalanx formation". The goal of this formation was to create uniformity and a powerful military force in order to maximize the effectiveness as the army as a whole, rather than use people as individual fighters. With the hoplite formation everyone was the same in battle. The Phalanx formation item of armor or equipment, thus 'hoplite' may approximate to 'armored man'. Hoplites were the citizen-soldiers of the Ancient Greek City-states. ... \\
        \underline{Ground Truth:} "A citizen-soldier", "militia groups", "Militiamen" \\
        \underline{\textbf{LLM Answer:}} According to the document, a hoplite was a citizen-soldier of Ancient Greece who was primarily armed with spears and shields. They were organized in battle into the "Phalanx formation" and were known for their uniform appearance, wearing the same armor and carrying the same weapons in the same position. \textbf{(\textcolor{darkgreen}{correct})} \\
        \hangindent=1cm \hangafter=0
        
        \hspace{-1cm}\textbf{Case Study 2} \\
        \textbf{Question:} What did Arthur Wynne invent? \\
        \textbf{Adaptive Judge Bar $\tau_q$}: \textbf{\texttt{-18.2506}} \\
        \underline{Filtered and Ordered Documents}: the "New York World" newspaper. He is best known for the invention of the crossword puzzle in 1913, when he was a resident of Cedar Grove, New Jersey. Wynne created the page of puzzles for the "Fun" section of the Sunday edition of the "New York World". For the December 21, 1913, edition, he introduced a puzzle with a diamond shape and a hollow center, the letters F-U-N already being filled in. He called it a "Word-Cross Puzzle." Although Wynne's invention was based on earlier puzzle forms, such as the word diamond, he introduced a number of innovations (e.g. the on January 14, 1945. On December 20, 2013, he was honored with an interactive Google Doodle commemorating the "100th anniversary of the first crossword puzzle" with a puzzle by Merl Reagle. ... \\
        \underline{Ground Truth:} "crossword puzzle", "Crossword", "Fill in crossword" \\
        \underline{\textbf{LLM Answer:}} According to the document, Arthur Wynne invented the crossword puzzle in 1913. \textbf{(\textcolor{darkgreen}{correct})} \\
        \hangindent=1cm \hangafter=0

        \hspace{-1cm}\textbf{Case Study 3} \\
        \textbf{Question:} The America's Cup trophy was held by America from 1852 until 1983 when the Cup was won by which challenger? \\
        \textbf{Adaptive Judge Bar $\tau_q$}: \textbf{\texttt{0.0222}} \\
        \underline{Filtered and Ordered Documents}: II won the Cup. 1983 America's Cup was the occasion of the first winning challenge to the New York Yacht Club, which had successfully defended the cup over a period of 132 years. An Australian syndicate representing the Royal Perth Yacht Club fielded the Australia II, skippered by John Bertrand against defender Liberty, skippered by Dennis Conner, won the match races to win the America's Cup, ending the longest winning streak in sporting history and ending U.S. domination of the racing series. ... \\
        \underline{Ground Truth:} "Australia II of Australia", "australia ii of australia" \\
        \underline{\textbf{LLM Answer:}} According to the document, the America's Cup trophy was held by the New York Yacht Club (NYYC) from 1857 until 1983. The Cup was won by an Australian syndicate representing the Royal Perth Yacht Club, fielding the Australia II, skippered by John Bertrand, in 1983. \textbf{(\textcolor{darkred}{wrong})} \\
        \hangindent=1cm \hangafter=0
    }
}
\caption{Case Study: \textbf{Adaptive Judge Bar $\tau_q$} (Dataset: \textbf{TriviaQA}; \texttt{LLM Agents}: \textbf{$\texttt{Llama3}_{\texttt{8B}}$})}
\label{apx:fig:case_trivia_llama3}
\end{figure*}

\end{document}